\pdfoutput=1
\PassOptionsToPackage{dvipsnames}{xcolor}
\PassOptionsToPackage{pdftex}{xcolor}
\documentclass[11pt]{article}

\usepackage[preprint]{acl}

\usepackage{times}
\usepackage{latexsym}
\usepackage{booktabs}
\usepackage{multirow}
\usepackage{amsmath} 
\usepackage{amsfonts}     
\usepackage{inconsolata}
\usepackage{microtype}
\usepackage{amsmath}
\usepackage{amssymb}
\usepackage{times}
\usepackage{latexsym}
\usepackage{algorithm}
\usepackage{algpseudocode}
\usepackage{times}
\usepackage{latexsym}
\usepackage{smartdiagram}
\usepackage{appendix}
\usepackage{tikz}
\usepackage{tcolorbox}
\usetikzlibrary{arrows.meta,
                positioning,calc
                }
\usepackage{amsmath}
      
\usepackage{hwemoji}
\usepackage{xcolor}
\usepackage{subcaption}

\usepackage{inconsolata}
\usepackage{amsthm, amsmath, amsfonts}
\usepackage{bm}
\usepackage{pgfplots}
\pgfplotsset{compat=1.17}
\definecolor{deepblue}{HTML}{1f77b4}
\definecolor{vibrantorange}{HTML}{ff7f0e}
\definecolor{lushgreen}{HTML}{2ca02c}
\definecolor{elegantpurple}{HTML}{9467bd}
\definecolor{olivegreen}{RGB}{128, 128, 0}
\definecolor{darkgreen}{RGB}{0, 100, 0}
\definecolor{limegreen}{RGB}{50,205,50}
\definecolor{salmon}{RGB}{250,128,114}
\definecolor{mistyrose}{RGB}{255,228,225}
\definecolor{colorA}{RGB}{26,112,149}    
\definecolor{colorB}{RGB}{241,171,64}  
\definecolor{colorC}{RGB}{120,175,92}   
\definecolor{colorD}{RGB}{151,65,113}

\tikzset{
  heading/.style={
    rectangle,
    rounded corners,
    draw=gray,
    align=center,
    minimum width=8em,
    text width=8em,
    font=\scriptsize,
    fill=yellow!50!red!20,
    top color=yellow!50!red!30,
    bottom color=yellow!50!red!10,
    drop shadow={shadow xshift=2pt, shadow yshift=-2pt, fill=gray!70, opacity=0.6}
  },
  example_heading/.style={
    rectangle,
    rounded corners,
    draw=gray,
    align=center,
    minimum width=8em,
    text width=10em,
    font=\scriptsize,
  },
  example_prompt/.style={
    rectangle,
    rounded corners,
    draw=gray,
    align=center,
    minimum width=8em,
    text width=40em,
    font=\scriptsize,
  },
  inline_prompt/.style={
    rectangle,
    rounded corners,
    draw=gray,
    align=left,
    minimum width=18em,
    text width=18em,
    font=\scriptsize,
  },
  prompt/.style={
    rectangle,
    rounded corners,
    draw=gray,
    align=left,
    minimum width=18em,
    text width=18em,
    font=\scriptsize,
    fill=green!60!blue!20,
    top color=green!70!blue!30,
    bottom color=green!70!blue!10,
    drop shadow={shadow xshift=2pt, shadow yshift=-2pt, fill=gray!70, opacity=0.6}
  },
  response/.style={
    rectangle,
    rounded corners,
    draw=gray,
    align=center,
    minimum width=10em,
    text width=18em,
    font=\scriptsize,
    fill=yellow!50!red!20,
    top color=yellow!50!red!30,
    bottom color=yellow!50!red!10,
    drop shadow={shadow xshift=2pt, shadow yshift=-2pt, fill=gray!70, opacity=0.6}
  },
  arrow/.style={
    ->,
    draw=gray,
    line width=2mm,
    shorten >=1pt,
    shorten <=1pt,
    -{Triangle[scale=0.6]}
  }
}

\usepackage{inconsolata}
\usepackage{arydshln}
\usepackage{multirow}
\usepackage[colorinlistoftodos,prependcaption,textsize=tiny]{todonotes}
\definecolor{olivegreen}{RGB}{107,142,35}
\definecolor{lightolivegreen}{RGB}{157,192,105}

\usepackage[T1]{fontenc}

\usepackage[utf8]{inputenc}

\usepackage{microtype}

\usepackage{inconsolata}

\usepackage{graphicx}	
\usepackage{xcolor}
\usepackage{soul}
\usepackage{lipsum}  

\usepackage{colortbl}
\usepackage{tabulary}
\usepackage{etoolbox}

\def \ourmodel{\texttt{Swan}}
\def \smallm{\texttt{Swan-Small}}
\def \largem{\texttt{Swan-Large}}
\def \benchmark{\texttt{ArabicMTEB}}
\def \benchl{\texttt{ArabicMTEB-Lite}}

\title{
\raisebox{-1.5ex}{\protect\includegraphics[height=3.9\fontcharht\font`\B]{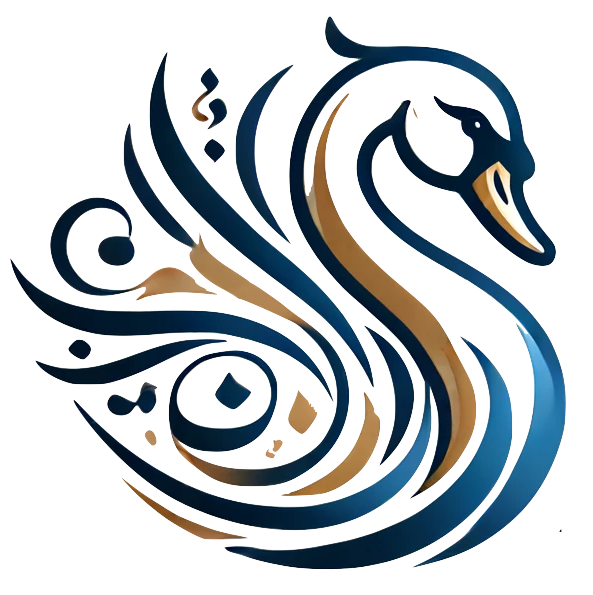}}
\ourmodel~\textbf{and}~\benchmark:\\[0.5ex]
Dialect-Aware, Arabic-Centric, Cross-Lingual, and Cross-Cultural Embedding Models and Benchmarks
}

\author{Gagan Bhatia~$^{\xi}$~El Moatez Billah Nagoudi~$^{\xi,\gamma}$ \bf{Abdellah El Mekki~$^{\omega}$ Fakhraddin Alwajih}~$^{\xi}$ \\ \bf{Muhammad Abdul-Mageed$^{\xi,\omega, \lambda}$}\\
\normalsize$^{\xi}$
  The University of British Columbia,~~~~~~\normalsize  $^{\omega}$ MBZUAI, ~~~~~~\normalsize $^{\gamma}$ PSU,  ~~~~~~\normalsize   $^{\lambda}$ Invertible AI\\ %
  \texttt{\normalsize \{gagan30@student.,muhammad.mageed@\}ubc.ca}}

\begin{document}
\maketitle
\begin{abstract}
We introduce {\bf \ourmodel}, a family of embedding models centered on Arabic, designed for both small-scale and large-scale applications. \ourmodel{} comprises two variants: \ourmodel-Small, built on ARBERTv2, and \ourmodel-Large, based on ArMistral, a pretrained Arabic large language model. To evaluate our models, we propose a comprehensive benchmark suite, dubbed \benchmark, that assesses cross-lingual, multi-dialectal, multi-domain, and multi-cultural Arabic text embedding performance. \benchmark covers eight diverse tasks sourced from 94 datasets. \ourmodel-Large achieves state-of-the-art results, outperforming Multilingual-E5-large in most Arabic tasks, while \ourmodel-Small consistently surpasses Multilingual-E5-base. Our extensive evaluations show that \ourmodel~models are both dialectally and culturally aware, achieving strong performance across diverse Arabic domains while maintaining significant cost efficiency. This work significantly advances the field of Arabic language modelling and provides valuable resources for future research and applications in Arabic NLP. Our models and benchmark are available at our GitHub page: \href{https://github.com/UBC-NLP/swan}{https://github.com/UBC-NLP/swan}.  



\end{abstract}

\section{Introduction}

NLP has seen rapid advancements in recent years, driven by groundbreaking developments in deep learning and the emergence of sophisticated distributed text representations such as word and sentence embeddings~\cite{devlin2018bert,reimers2019sentence}. These embeddings, which transform text into dense vectors, enable effective semantic understanding and are pivotal for enhancing performance across many downstream applications, including text classification, semantic search, and machine translation. Moreover, text embeddings have become fundamental to the success of large language models (LLMs)~\cite{touvron2023llama,jiang2023mistral7b,gemmateam2024gemma},
which are increasingly integrated into a variety of real-world systems and tools. One of the most promising applications of these embeddings is in the realm of Retrieval-Augmented Generation (RAG)~\cite{shao2023enhancing,rag2023domain}, where LLMs are augmented with information retrieval capabilities. In RAG-based systems, lightweight embedding models retrieve relevant information from large corpora, fed as context to models like ChatGPT~\cite{openai2023chatgpt} or GPT-4~\cite{openai2024gpt4}. This synergy between embeddings and LLMs has demonstrated significant improvements in both general-purpose tasks such as question answering~\cite{lin2023li,rag2023domain} and domain-specific applications~\cite{bhatia2024fintral,shi2023raft,lin2023li}.
\begin{figure}[t]
\includegraphics[width=\columnwidth]{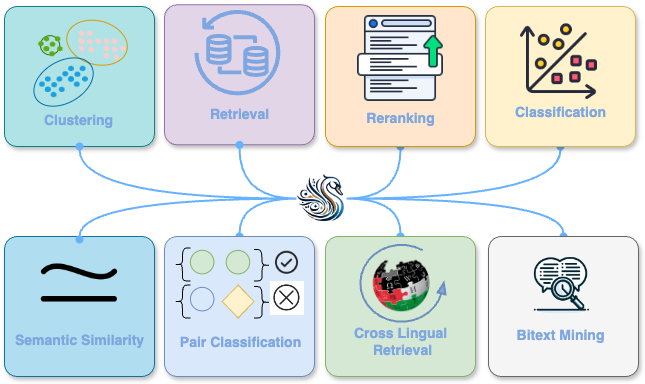}
\centering
\caption{Overview of our \benchmark~benchmark tasks, covering \textit{clustering, retrieval, reranking, classification, semantic similarity, pair classification, cross-lingual retrieval,} and \textit{bitext mining}.
}
\label{fig:Arabic MTEB}
\end{figure}

Despite these advances, the predominant focus of current embedding models has been on English and Chinese, which limits their applicability to other languages. This is particularly true for Arabic, a collection of languages, language varieties, and diverse dialects with rich morphology~\cite{abdul2023nadi,abdul2024nadi}, making it challenging to develop effective language representations~\cite{nagoudi-etal-2022-turjuman,huang-etal-2024-acegpt}. Existing multilingual models often fail to capture these complexities, leading to a suboptimal performance on Arabic NLP tasks \cite{abdul-mageed2020arbert,elmadany2022orca}. Addressing this limitation requires the development of Arabic-specific embedding models that are sensitive to the linguistic and cultural nuances of Arabic.

In this work, we introduce \ourmodel, a family of dialect-aware, Arabic-centric, cross-lingual, and cross-cultural embedding models designed to bridge this gap and push the boundaries of Arabic NLP. Our contributions are as follows:\textbf{(1)}~We introduce \ourmodel, a cutting-edge family of Arabic embedding models. This includes two variants: \smallm, based on ARBERTv2~\cite{elmadany2022orca}, and \largem, built upon ArMistral, a further pretrained Arabic language model. 
\textbf{(2)}~We present \benchmark, a comprehensive evaluation benchmark for Arabic text. \benchmark~is designed to assess cross-lingual, multi-dialectal, multi-domain, and multi-cultural performance, spanning eight tasks and $94$ datasets.  Figure~\ref{fig:Arabic MTEB} provides an overview of \benchmark.
\textbf{(3)}~Our larger model, \largem, showcases state-of-the-art text embedding capabilities, surpassing Multilingual-E5-large~\cite{wang2024multilingual} on most Arabic tasks. Similarly, our smaller, \smallm, consistently outperforms Multilingual-E5-base~\cite{wang2024multilingual} on most Arabic tasks. 
\textbf{(4)}~Through rigorous benchmarking, we demonstrate that \ourmodel~models are not only dialectally and culturally aware, but also excel across diverse Arabic domains while maintaining a significantly lower monetary cost. 

The rest of the paper is organized as follows: In Section~\ref{section: related_works}, we review related work with a particular emphasis on Arabic text embedding models, their applications and challenges. We present our approach to model training of~\ourmodel~models in Section~\ref{section: models}. Section~\ref{section: benchmark} outlines how we built our benchmark dataset,~\benchmark. Section~\ref{section: evaluation} is about our experiments and model analysis. We conclude in Section~\ref{section: conclusion}.

\section{Related Work} \label{section: related_works}
\begin{table*}[!htp]\centering
\scriptsize
\resizebox{\textwidth}{!}{%
\begin{tabular}{lllcccc}\toprule
\textbf{Benchmark } &\textbf{Lang } &\textbf{Tasks} &\textbf{Datasets} &\textbf{Tasks} &\textbf{CRTR} &\textbf{Ar Cul/Dom} \\\midrule
MTEB \cite{muennighoff2022mteb} &English &\textit{RTR, STS, PairCLF, CLF, RRK, CLR, SUM} &56 &7 &$\times$ &$\times$ \\
C-MTEB \cite{xiao2023c-pack} &Chinese &\textit{RTR, STS, PairCLF, CLF, RRK, CLR} &35 &6 &$\times$ &$\times$ \\
De-MTEB \cite{sturua2024jinaembeddingsv3multilingualembeddingstask}&German &\textit{RTR, STS, PairCLF, CLF, RRK, CLR} &17 &6 &$\times$ &$\times$ \\
F-MTEB \cite{ciancone2024extending} &French &\textit{RTR, STS, PairCLF, CLF, RRK, CLR, BTM}&17 &7 &$\times$ &$\times$ \\
Es-MTEB\cite{mohr2024multi-task} &Spanish &\textit{RTR, STS, PairCLF, CLF, RRK, CLR }&17 &6 &$\times$ &$\times$ \\
Polish \cite{poswiata2024pl} &Polish &\textit{RTR, STS, PairCLF, CLF, CLR} &26 &5 &$\times$ &$\times$ \\
Ru-MTEB \cite{poswiata2024pl} &Russian &\textit{RTR, STS, PairCLF, CLF, RRK, CLR} &23 &6 &$\times$ &$\times$ \\
\multirow{3}{*}{Scand. \cite{enevoldsen2024scandinavian}} &Danish &\multirow{3}{*}{\textit{RTR, CLF, BTM, CLR}} &\multirow{3}{*}{26} &\multirow{3}{*}{4} &$\times$ &$\times$ \\
&Norweg. & & & &$\times$ &$\times$ \\
&Swedish & & & &$\times$ &$\times$ \\\midrule
\benchmark~(Ours) &Arabic &\textit{RTR, STS, PairCLF, CLF, RRK, CLR, BTM, CRTR }&94 &8 &\checkmark &\checkmark \\
\bottomrule
\end{tabular}}
\caption{Comparison of various text embedding benchmarks proposed in the literature across the different covered task clusters. \textbf{RTR}: retrieval, \textbf{STS}: semantic textual similarity, \textbf{PairCLF}: pair classification, \textbf{CLF}: classification, \textbf{CLR}: clustering, \textbf{RRK}: reranking, \textbf{BTM}: bitext mining, \textbf{CRTR}: cross-lingual retrieval. }\label{tab: MTEB_comp}
\end{table*}

In recent years, there have been remarkable advancements in text embedding models, with a shift towards developing universal embeddings for diverse tasks and domains. Despite this, specialized models and benchmarks for languages like Arabic remain underexplored.


\noindent\textbf{Multilingual Text Embedding Models.} With the need for language-agnostic embeddings growing, multilingual models such as LASER \cite{artetxe-schwenk-2019-massively} and LaBSE \cite{feng-etal-2022-language} were developed using BiLSTM and Transformer encoders, respectively. Building on this, the Multilingual-E5 \cite{wang2024multilinguale5textembeddings} series extends the E5 architecture to support diverse languages using multilingual text pairs and synthetic data. GRIT \cite{muennighoff2024generativerepresentationalinstructiontuning} further unifies generative and embedding tasks within a single model. Newer models such as ColBERT-XM~\cite{louis2024colbert-xm} and Gecko~\cite{lee2024gecko} refine multilingual embeddings through modular and distilled architectures.

\noindent\textbf{Arabic-Specific Models.} Despite progress in Arabic NLP, existing models are not optimized for Arabic text embedding and retrieval. Efforts like ARBERT~\cite{abdul-mageed-etal-2021-arbert} and AraMus \cite{alghamdi-etal-2023-aramus} have focused on encoding and generation but are not tailored for sentence-level embeddings. While language-agnostic models such as LASER and Multilingual-E5 include Arabic in their training data, they may not fully capture its linguistic intricacies and diversity. To address this,~\citet{nacar2024enhancing} introduced models and training datasets to improve semantic similarity performance for Arabic.

\noindent\textbf{Text Embedding Benchmarks.} Most text embedding evaluations rely on a narrow set of datasets, limiting their generalisation ability. To address this, the Massive Text Embedding Benchmark (MTEB) \cite{muennighoff-etal-2023-mteb} introduced eight task categories with 58 datasets and 112 languages. However, it remains predominantly focused on English. Similar benchmarks have been developed for other languages, such as C-MTEB \cite{xiao2023c-pack} for Chinese. For Arabic, evaluations have primarily centred on Semantic Text Similarity (STS) tasks \cite{nacar2024enhancing}. However, excelling in STS does not guarantee optimal performance in tasks like clustering or reranking \cite{muennighoff-etal-2023-mteb}. Existing Arabic benchmarks like ORCA \cite{elmadany-etal-2023-orca} and ALUE \cite{seelawi-etal-2021-alue} focus on natural language understanding (NLU), while Dolphin~\cite{nagoudi-etal-2023-dolphin} targets natural language generation (NLG). This work is the first comprehensive benchmark for evaluating Arabic text embeddings across multiple tasks.

\section{Swan } \label{section: models}
\subsection{Training Data} \label{subsection: training_data}

To train {\ourmodel}, we develop the most extensive training corpus for Arabic embedding models, leveraging a unique assembly of datasets to ensure comprehensive linguistic coverage and diversity. Our training data covers paragraph-based and sentence-based datasets curated from multiple sources. \autoref{tab: train_data} shows an overview of our training datasets. 


\begin{figure}




        \centering
        \begin{tikzpicture}[node distance=12mm]
	\node (n1) [prompt] {
\textbf{You have been assigned a retrieval task: \{task\}} \\
Your mission is to write one text retrieval example for this task in JSON format. \\
The JSON object must contain the following keys: \\
\phantom{new} \textit{user\_query}: a string, a query specified by the retrieval task. \\
\phantom{new} \textit{positive}: a string, a relevant document for the user query. \\
\phantom{new} \textit{hard\_negative}: a string, a document closely related to the query. \\
\textbf{Please adhere to the following guidelines:} \\
The user\_query should be paragraph-based, understandable with some effort or ambiguity, and diverse in topic. The hard\_negative contains some useful information, but it should be less useful or comprehensive than the positive.  \\
}; 

\node (n3) [response, below=of n1]
	      {
\includegraphics[width=\textwidth]{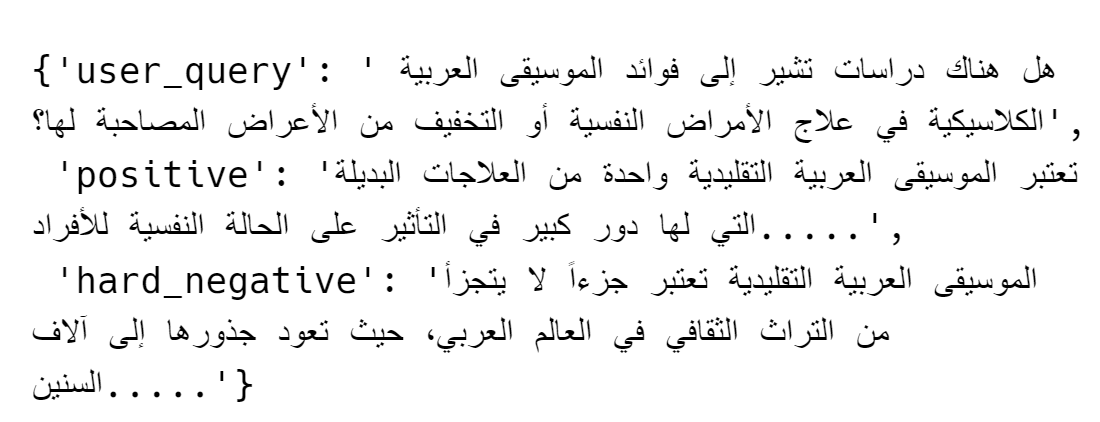}
};
\draw [arrow] (n1) --  (n3);
\node[above left=0.15cm and -3.25cm of n3] (image) {\includegraphics[height=0.7cm]{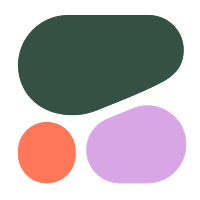}};
\node[below= -0.25cm of image] {\scriptsize Cohere};
\end{tikzpicture}
    \caption{Methodology to generate our synthetic data.}
    \label{fig:display_data_gen}
\end{figure}

\noindent \noindent\textbf{MSA Datasets.} We focus on two sources: \textbf{(i) Human-generated data:} Composed from ORCA~\cite{elmadany-etal-2023-orca} and mMARCO~\cite{bonifacio2021mmarco}. ORCA is a compilation of labelled datasets with tasks such as semantic text similarity (STS), sentence classification, text classification, natural language inference (NLI), and question answering. We use all the training sets from ORCA, encompassing $60$ different datasets. mMARCO-ar is the translated version of MARCO, which is a human-generated dataset \cite{bajaj2018msmarcohumangenerated}. Both of these datasets are cleaned up and de-duplicated using Polydedupe ~\cite{Gagan_PolyDeDupe_2023},\footnote{\href{https://github.com/gagan3012/PolyDeDupe}{https://github.com/gagan3012/PolyDeDupe}} which is further described in Appendix \ref{sec:pipeline}. \textbf{ (ii) Synthetically-generated data:} To augment our MSA training data for retrieval tasks, we use Command R+ \cite{cohere_for_ai_2024} to generate high-quality synthetic data.\footnote{We performed various in-house evaluations comparing multiple models. Command R+ was chosen as it is open-source and efficient in generating Arabic varieties (standard and dialectal).} The generation methodology is inspired by \citet{wang2024improvingtextembeddingslarge}, and we employ the procedure shown in \autoref{fig:display_data_gen} to generate our synthetic dataset. We generate $100k$ in general MSA data and $5k$ in instances for specific domains such as finance, news, medicine, and legal for a total of $120k$ MSA instances. 
\begin{table}[t]\centering
\scriptsize
\resizebox{\columnwidth}{!}{%
\begin{tabular}{llllllr}\toprule
\textbf{Family} &\textbf{Language} &\textbf{Type} &\textbf{Dataset} &\textbf{Level} &\textbf{Size} \\\midrule
\multirow{6}{*}{\textbf{Monoling}} &\multirow{6}{*}{Ar} &\multirow{3}{*}{Human} &ORCA-MSA &\multirow{3}{*}{Sent} &378K \\
& & &ORCA-DIA & &122K \\
& & &MMARCO-ar & &8.1M \\
& &\multirow{3}{*}{Synthetic} &Synth-MSA &\multirow{3}{*}{Parag} &100K \\
& & &Synth-DIA & &15K \\
& & &Synth-DOM & &20K \\\midrule
\multirow{2}{*}{\textbf{Crossling}} &Ar to 15 lg &\multirow{2}{*}{Human} &MMARCO &\multirow{2}{*}{Sent} &3M \\
&Ar to 6 lg & &XOR-TyDi & &20.5K \\\midrule
\multirow{2}{*}{\textbf{Multiling}} &11 lg & \multirow{2}{*}{Human}&Mr-Tydi & \multirow{2}{*}{Sent}&49K \\
&16 lg & &Miracl & &343K \\\midrule
\multicolumn{5}{l}{\textbf{Total}} &12.5M \\
\bottomrule
\end{tabular}}
\caption{The diverse datasets employed for training our Arabic embedding models. In the synthetic dataset, we have three datasets: the MSA dataset, the dialectal dataset (Egyptian and Moroccan), and domain-based focusing on medical, financial, legal and news domains.}\label{tab: train_data}
\end{table}

\noindent \noindent\textbf{Dialectal Arabic Datasets.} Similar to the MSA datasets, we focus on two sources: \textbf{(i) Human-generated data:} We use publicly available dialectal Arabic data, which primarily covers Gulf, Egyptian, Moroccan, and Levantine varieties of Arabic \cite{elmadany2022orca,nagoudi2023dolphin,alwajih2024dallahdialectawaremultimodallarge,abdul-mageed-etal-2020-nadi,abdul-mageed-etal-2021-nadi,abdul-mageed-etal-2022-nadi,abdul-mageed-etal-2023-nadi,abdul-mageed-etal-2018-tweet,abdul-mageed-etal-2020-toward,keleg-etal-2023-aldi,keleg-magdy-2023-arabic,zaidan-callison-burch-2014-arabic,bouamor-etal-2018-madar}. The total number of samples is $122K$. \textbf{(ii) Synthetically-generated data:} As most human-generated dialectal data comes from noisy environments such as social media, it often results in short texts of low quality. Thus, we use Command-R+ to generate paragraph-based synthetic data for Egyptian and Moroccan dialects to improve the performance of our models on dialectal Arabic. We generated $15k$ dialectal instances using the same methodology as our synthetic MSA datasets described above. 

\noindent \noindent\textbf{Cross-Lingual \& Multilingual Datasets.} To adapt our model for cross-lingual and multilingual scenarios, we incorporate the mMARCO dataset, which provides translations of the MS MARCO dataset into 15 languages~\cite{bonifacio2021mmarco}. To ensure that documents correspond accurately to their queries in different languages, we utilize specific IDs. We create $100k$ samples for each cross-lingual pair and shuffle the IDs to prevent repetition, thus guaranteeing that unique data samples are used for each language. We utilize the MIRACL~\cite{zhang2022making}, XOR-TyDI~\cite{asai2021xorqacrosslingualopenretrieval}, and Mr.TyDi~\cite{zhang-etal-2021-mr} datasets as our crosslingual and multilingual resources.

\subsection{Training Strategy}
For \ourmodel, we consider two models: \smallm~ and \largem.
The choice of training two models with different sizes is driven by the need to balance performance and computational efficiency. \smallm~is designed to cater to scenarios where lower computational resources are available or when a lightweight model is preferred for deployment on edge devices. In contrast, \largem~is intended for settings where achieving SoTA performance is paramount, leveraging a larger parameter size to better capture the nuances of Arabic. 


\noindent\textbf{Data Preprocessing.} We incorporate human-generated and synthetic datasets into our training pipeline to ensure robust performance across various dialects and cultural contexts. We first train on MSA datasets, followedby fine-tuning on dialectal datasets. This two-step approach ensures that both MSA and dialectal varieties are well represented, promoting better generalization across the full spectrum of Arabic varieties. Our dataset is constructed with a query format, including positive and negative samples.

\noindent\textbf{\smallm.} Built upon ARBERTv2 \cite{abdul2020arbert}, a powerful BERT-based model for the Arabic language. Here, our model is trained using the InfoNCE loss \cite{oord2019representationlearningcontrastivepredictive}, which maximizes the similarity between related sentences while minimizing the similarity between unrelated sentences. The model is trained for five epochs on the entire dataset with a learning rate of $5e^{-6}$ and a batch size of 128, incorporating 15 hard negatives. \smallm~has $164$M parameters and a dimension size of $768$. 

\noindent\textbf{\largem.} \largem~is based on ArMistral-7B, an in-house further pretrained version of Mistral-7B \cite{jiang2023mistral7b}\footnote{Further details about ArMistral can be found in Appendix~\ref{Appendix: training_mistral}.}. To train \largem, we use LoRA \cite{hu2021lora} for parameter efficient training and InfoNCE loss for optimization. We train the model for three epochs on the entire dataset with a learning rate of $5e^{-6}$ and a batch size of $128$, incorporating seven hard negatives. \largem~has $7.2B$ parameters and a dimension size of $4,096$. 

\subsection{Training Methodology} \label{Appendix: training}

Given a relevant query-document pair \((q^+, d^+)\), we modify the query by appending an instructional template to it. This process transforms the original query \(q^+\) into a new form \(q^+_\text{inst}\) as defined below:

\begin{equation*} \label{equ:instruction_template}
    q^+_\text{inst} = \text{Instruction: \{task\_instruction\} Query:} \{q^+\}
\end{equation*}

Here, ``\emph{\{task\_instruction\}}'' refers to a one-sentence description of the embedding task taken from Table \ref{tab: defs}, which outlines the instructions for different tasks. Using a pretrained LLM, we append an \textit{[EOS]} token at the end of both the modified query and the document. These are then fed into the LLM to extract embeddings \(\mathbf{h}_{q^+_\text{inst}}\) and \(\mathbf{h}_{d^+}\) from the vector at the last [EOS] layer. Again, training of the embedding model is conducted using the InfoNCE loss function, which is widely recognized for its effectiveness in learning high-quality embeddings. The objective is minimized using the following formulation:

\begin{equation*} \label{equ:infonce}
   \min \left(-\log \frac{\phi(q^+_\text{inst}, d^+)}{\phi(q^+_\text{inst}, d^+) + \sum_{n_i \in \mathbb{N}} \phi(q^+_\text{inst}, n_i)}\right)
\end{equation*}

In the equation above, \(\mathbb{N}\) denotes the set of negative samples, and \(\phi(q, d)\) is the similarity scoring function between a query \(q\) and a document \(d\).

\subsection{Inclusion of Hard-Negatives}
To enhance the model's performance, it is crucial to use negative documents closely aligned with the query's context \cite{karpukhin2020dense,khondaker2022benchmark}. 
This method allows us to observe the impact of introducing more challenging or "hard" negatives into the training process. We only generate hard negatives for the Arabic subset of our training data from Section~\ref{subsection: training_data}. We found that using $15$ hard negatives for \smallm~yields the best performance, whereas for our bigger model, \largem, the model overfits a more significant number of hard negatives, and $7$ gives us the best performance.  

\noindent \textbf{Impact of Hard Negatives.} Hard negatives in contrastive learning are examples that closely resemble correct instances but are ultimately incorrect. Their inclusion encourages the model to learn finer-grained distinctions, improving its ability to differentiate between similar but distinct classes. The process involves converting all documents into a vector form within the embedding space. Subsequently, these document embeddings are compared using the cosine similarity score to establish their relevance to the query. Once all documents are scored, they are sorted by their similarity to the query. The top-ranked document is typically the positive example, while the rest are potential negatives. Our experiments assess the impact of varying the hard negatives used while training our models, \largem~and \smallm. We train each model with different quantities of hard negatives. Namely, we experiment with using hard negatives values from the set \{\textit{1, 3, 7, 15, 31}\} per training instance. 
\noindent\smallm~achieves its highest performance of 56.25 with 15 hard negatives. The model exhibits a general upward trend as the number of hard negatives increases, peaking at 15 before slightly declining at 31. This pattern suggests that while additional hard negatives initially enhance learning by introducing valuable challenges, excessive complexity may lead to diminishing returns, ultimately hindering further improvement.
\noindent\largem~achieves its peak performance of 60.42 when trained with seven hard negatives, suggesting an optimal balance that enhances learning without overloading the model. Notably, increasing the number of hard negatives beyond this point does not lead to further gains, indicating a threshold where additional complexity ceases to improve learning outcomes.
\begin{table}[!t]\centering
\resizebox{\columnwidth}{!}{%
\begin{tabular}{lccccc}\toprule
\textbf{Model (HN)} &\textbf{1} &\textbf{3} &\textbf{7} &\textbf{15} &\textbf{31} \\\midrule
\cellcolor{cyan!0}\textbf{\smallm} &\cellcolor{cyan!0}48.84 &\cellcolor{cyan!0}52.19 &\cellcolor{cyan!0}\underline{54.13} &\cellcolor{cyan!0}\textbf{56.25} &\cellcolor{cyan!0}51.93 \\\midrule
\cellcolor{green!0} \textbf{\largem} &\cellcolor{green!0} 59.48 &\cellcolor{green!0} 59.35 &\cellcolor{green!0} \textbf{60.42} &\cellcolor{green!0}59.44 &\cellcolor{green!0} \underline{59.83} \\
\bottomrule
\end{tabular}}
\caption{Impact of number of Hard Negatives (HN).}\label{tab: hard_negs}
\end{table}





\section{ArabicMTEB Benchmark} \label{section: benchmark}




\begin{table}[t]\centering
\scriptsize
\resizebox{\columnwidth}{!}{%
\begin{tabular}{lcccl}\toprule
\textbf{Task} &\textbf{Datasets} &\textbf{Langs} &\textbf{Dialects} &\textbf{Metric} \\\midrule
\textbf{RTR} &36 &1 &4 &nDCG@10 \\
\textbf{CRTR} &12 &7 &0 &nDCG@10 \\
\textbf{CLF} &18 &1 &6 &AP \\
\textbf{BTM} &11 &5 &8 &F1 \\
\textbf{RRK} &5 &2 &0 &MAP \\
\textbf{STS} &5 &1 &3 &Spearman Corr \\
\textbf{CLR} &4 &1 &0 &v-measure \\
\textbf{PairCLF} &3 &1 &0 &AP \\\bottomrule
\textbf{Total} &\textbf{94} &\textbf{9} &\textbf{11} & \\
\bottomrule
\end{tabular}}
\caption{Overview of our Tasks in \benchmark. \textbf{$^{*}$}Total represents the unique languages.}\label{tab: overview}
\end{table}
In this section, we present \benchmark, a comprehensive Arabic-centric text embedding benchmark designed to evaluate text embeddings across a wide range of tasks and scenarios. \benchmark~addresses the limitations of existing benchmarks that either exclude Arabic or lack coverage of diverse Arabic language varieties, dialects, and cultural nuances. Our benchmark includes $94$ datasets spanning $8$ distinct tasks, as summarized in Table~\ref{tab: overview}. Further details about the datasets used in \ benchmark~ can be found in Appendix \ref{sec:appendix_datasets_benchmark}. 
\benchmark~was developed to provide comprehensive coverage of Arabic text embedding capabilities, ensuring the inclusion of MSA and other varieties. It offers diverse task types, such as retrieval, classification, and semantic similarity, to evaluate embeddings holistically across different scenarios by incorporating novel domain-specific, dialectal, and country-level culturally aware datasets, \benchmark~represents a more applicable and realistic assessment of Arabic text embeddings.

\subsection{Task Categories}
\benchmark~categorizes evaluation datasets into the following key task categories, with each type providing a unique perspective on the capabilities of text embeddings. The corresponding metadata for each task, covering the considered number of datasets, number of languages, number of dialects, and evaluation metric, is presented in Table~\ref{tab: overview}.

\noindent\textbf{Arabic Text Retrieval.} This task uses Arabic queries to retrieve Top-$k$ relevant documents from a large Arabic corpus. \benchmark~includes $35$ retrieval datasets such as XPDA~\cite{shen2023xpqa} and Dolphin's long-form QA datasets~\cite{nagoudi2023dolphin}. Including these datasets helps evaluate complex information retrieval scenarios in Arabic.

\noindent\textbf{Bitext Mining.} This task identifies sentence-level translations between different languages and dialects. \benchmark~includes $12$ datasets spanning various language pairs like Arabic to French and English. This task is crucial for understanding text embeddings' cross-lingual and dialectal translation capabilities.

\noindent\textbf{Cross-Lingual Retrieval.} This task uses Arabic queries to retrieve documents in other languages, such as English, German, Spanish, and Chinese. \benchmark~employs the mMarco Dev set~\cite{bonifacio2021mmarco} and includes $11$ language pairs. 

\noindent\textbf{Re-Ranking.} This task reorders candidate documents for a query based on embedding similarity scores. \benchmark~features five re-ranking datasets such as MIRACL~\cite{zhang-etal-2023-miracl}, enabling the evaluation of embeddings' ability to refine search results. 

\noindent\textbf{Semantic Textual Similarity (STS).} STS measures the correlation between the embeddings of two sentences, assessing their semantic similarity. \benchmark~includes five STS datasets like STS17 and STS22~\cite{cer-etal-2017-semeval}, along with two synthetic datasets generated using GPT-4 (Details of the creation of these datasets can be found in Appendix \ref{sec:sts}). 

\noindent\textbf{Classification.} Classification predicts labels for input texts using text embeddings. \benchmark~comprises $18$ multi-domain datasets, from ORCA~\cite{elmadany2022orca}. This task evaluates models' ability to categorize Arabic text accurately, making it a valuable benchmark for downstream tasks such as sentiment analysis.

\noindent\textbf{Pair Classification.} This task predicts the relationship between two sentences based on their embeddings. \benchmark~includes three datasets, such as XNLI~\cite{conneau2018xnli}. 

\noindent\textbf{Clustering.} Clustering groups sentences into clusters based on embedding similarity, evaluating unsupervised learning performance. \benchmark~includes four clustering datasets, such as Arabic News Articles and stance detection datasets from ~\cite{baly2018integrating}. 

\subsection{Dialectal ArabicMTEB}

Dialectal \benchmark~is a specialized fork of the original \benchmark, focusing exclusively on Arabic dialectal datasets. This extension addresses the unique challenges posed by the significant variations in Arabic dialects across different regions, which have been underrepresented in NLP research.
While research on dialectal Arabic text embedding has been limited, dialectal \benchmark~fills this gap by providing a comprehensive collection of $19$ datasets specifically curated to evaluate embeddings' performance on diverse Arabic dialects. These datasets span multiple tasks, offering a robust framework for assessing model performance across various dialectal contexts:  \noindent\textbf{(1) Bitext Mining.} Eight datasets covering dialects such as Algerian, Egyptian, Jordanian, Lebanese, Moroccan, Saudi, and Yemeni~\cite{nagoudi-etal-2022-turjuman, bouamor-etal-2014-multidialectal}.
\textbf{(2) Retrieval.} Five datasets focusing on dialects from Algeria, Egypt, Morocco, and the Gulf regions~\cite{nagoudi2023dolphin}.
\textbf{(3) Classification:} Five datasets for binary, regional, and country-level dialect identification~\cite{mageed-etal-2021-nadi, abdul-mageed-etal-2024-nadi, elmadany2022orca, abdul2020arbert, ahmed-etal-2024-alclam}.
\textbf{(4) STS.} A novel synthetic dataset for Egyptian text similarity generated using Command-R+.

\subsection{Domain-Specific ArabicMTEB}




Arabic Text retrieval tasks are currently trending in real-world applications. They are utilized across multiple fields, including healthcare, finance, and legal sectors. Having specialized evaluation datasets is crucial for building text embeddings tailored to these domains. To meet this need, we introduce domain-specific \benchmark, a specialized fork of the broader \benchmark~benchmark. Domain-specific \benchmark~focuses on the news, finance, legal, medical, and general knowledge domains, offering a closer approximation to real-world scenarios. The creation of this benchmark involves collecting Arabic documents from these specialized sources and from Arabic Wikipedia. We then segment and chunk the documents into texts of $1,024$ tokens each. Subsequently, we randomly select chunks and employ GPT4-Turbo~\cite{openai2024gpt4} to generate five different styles of queries for each chunk. We filter out duplicate and repeated queries using GPT3.5 \cite{openai2024gpt4} to ensure a high-quality evaluation dataset. Our evaluation data creation pipeline is visualized in \autoref{fig:display_data_gen}.
The resulting benchmark, which we call \benchl, contains $10k$ queries and $100k$ documents spanning the domains described above.


\begin{figure}[t]
\includegraphics[width=1\columnwidth]{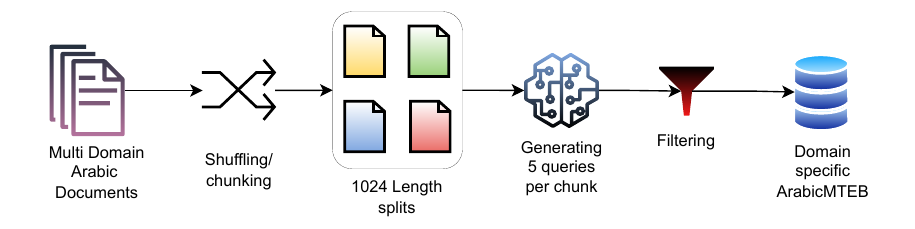}
\centering
\caption{Generation pipeline for our domain specific~\benchmark.}
\label{fig:air}
\end{figure}




     

\subsection{Cultural ArabicMTEB}

To show that our models are culturally aware, we have introduced Cultural \benchmark, a collection of datasets from 20 different Arab countries where we focus on specific cultural aspects like their geography, history, etc.
To construct the Cultural \benchmark, we use Arabic Wikipedia as our primary data source. For each included Arab country, we extract articles related to that country from its corresponding Wikipedia portal. The portal covers multiple categories (e.g., geography, economy, history) with subcategories (e.g., local movies, food items). This process resulted in $5$K to $55$K articles per country. Next, we generate retrieval questions and passages for each country. For this, we use GPT-4o-mini to develop, for each passage (from an article), a corresponding question whose specific answer is available within the passage itself. Following the same methodology, but applied to Egyptian and Moroccan dialectal versions of Wikipedia, we generate dialectal queries and their corresponding passages using Command-R+. Cultural \benchmark~ contains $1k$ queries and an average of $15k$ documents from various countries as described above.

\section{Evaluation} \label{section: evaluation}
\begin{table*}[h]\centering
\scriptsize
\resizebox{\textwidth}{!}{%
\begin{tabular}{lrrcccccccc}\toprule
\textbf{Model} &\textbf{Size~} &\textbf{Dim.~} &\textbf{RTR} &\textbf{STS} &\textbf{PairCLF} &\textbf{CLF} &\textbf{RRK} &\textbf{CLR} &\textbf{BTM} &\textbf{Avg.} \\\midrule
Arabert-v2-base &160M &768 &8.62 &39.77 &66.30 &55.77 &60.03 &41.74 &0.70 &38.99 \\
CamelBERT &163M &768 &9.21 &47.69 &67.43 &55.66 &60.20 &39.89 &1.85 &40.28 \\
ARBERTv2 &164M &768 &15.12 &47.88 &68.87 &\underline{56.85} &62.21 &39.25 &1.99 &41.74 \\
ATM-V2 &135M &768 &37.45 &55.90 &70.12 &46.42 &61.45 &32.35 &12.98 &45.24 \\
text2vec &118M &384 &27.69 &\textbf{59.37} &71.41 &47.94 &57.76 &37.26 &38.32 &48.54 \\
LaBSE &471M &768 &34.98 &54.15 &70.60 &49.57 &62.17 &41.42 &33.28 &49.45 \\
Me5-small &118M &384 &55.14 &56.73 &73.97 &50.85 &67.92 &\underline{42.37} &\underline{38.47} &55.06 \\
Me5-base &278M &768 &\underline{56.91} &57.99 &\underline{74.30} &52.30 &\textbf{69.07} &\textbf{42.56} &33.90 &\underline{55.29} \\
\midrule
\textbf{\smallm} &164M &768 &\textbf{58.42} &\underline{59.34} &\textbf{74.93} &\textbf{57.34} &\underline{68.43} &40.43 &\textbf{42.45} &\textbf{57.33} \\\midrule
e5-mistral-7b &7110M &4096 &56.34 &57.02 &70.24 &53.21 &66.24 &39.44 &\underline{70.5} &59.00 \\
Me5-large &560M &1024 &\underline{64.01} &\textbf{59.45} &\underline{75.06} &\underline{53.43} &\textbf{70.79} &\textbf{42.49} &66.33 &\underline{61.65} \\\midrule
\textbf{\largem} &7230M &4096 &\textbf{65.63} &\underline{59.10} &\textbf{75.62} &\textbf{54.89} &\underline{69.42} &\underline{41.24} &\textbf{71.24} &\textbf{62.45} \\
\bottomrule
\end{tabular}}
\caption{Overall {\benchmark} results. }\label{tab: results}
\end{table*}
\begin{table}[!htp]\centering
\scriptsize
\resizebox{\columnwidth}{!}{%
\begin{tabular}{lrrrrr}\toprule
\textbf{Model} &\textbf{RTR} &\textbf{STS} &\textbf{CLF} &\textbf{BTM} &\textbf{Avg.} \\\midrule
Arabert-v2-b &8.67 &41.64 &47.97 &0.99 &24.82 \\
MARBERT &5.45 &50.06 &53.46 &2.34 &27.83 \\
ARBERTv2 &7.52 &49.36 &54.31 &2.51 &28.43 \\
CamelBERT &6.92 &59.48 &50.69 &2.65 &29.93 \\
AlcLaM &8.56 &50.90 &\textbf{54.74} &7.54 &30.44 \\
ATM-V2 &36.23 &74.13 &34.39 &11.67 &39.10 \\
Me5-base &61.60 &74.84 &34.87 &3.30 &43.65 \\
Me5-small &57.61 &76.35 &34.78 &12.35 &45.27 \\
Me5-large &66.88 &77.02 &35.47 &51.08 &57.61 \\
e5-mistral-7b &\underline{72.35} &\underline{77.37} &35.91 &57.62 &60.81 \\\midrule
\textbf{\smallm} &63.16 &76.57 &\underline{54.52} &\underline{59.38} &\underline{63.41} \\
\textbf{\largem} &\textbf{77.03} &\textbf{79.22} &53.46 &\textbf{72.10} &\textbf{70.45} \\
\bottomrule
\end{tabular}}
\caption{Dialectal {\benchmark} results.}\label{tab: ArabicDialect}
\end{table}
\begin{table}[!htp]\centering
\resizebox{\columnwidth}{!}{%
\begin{tabular}{lccccccc}\toprule
\textbf{Model} &\textbf{News} &\textbf{Legal} &\textbf{Med} &\textbf{Fin} &\textbf{Wiki} &\textbf{Avg} &\textbf{Cost} $\downarrow$ \\\midrule
\textbf{\largem} &\textbf{90.42} &\textbf{89.96} &\textbf{81.64} &\underline{57.34} &\textbf{93.10} &\textbf{82.49} &\underline{0.75\$} \\
Openai-3-lg &\underline{88.1} &\underline{89.68} &\underline{80.24} &\textbf{61.46} &\underline{91.52} &\underline{82.20} &9.88\$ \\
Cohere-v3.0 &85.23 &86.52 &63.27 &42.80 &90.96 &73.76 &7.54\$ \\\midrule
\textbf{\smallm} &81.55 &78.86 &70.97 &42.48 &80.46 &70.86 &\textbf{0.44\$} \\
Openai-3-small &71.42 &85.23 &71.50 &32.90 &82.20 &68.65 &3.75\$ \\
Cohere-light-v3.0 &70.32 &86.83 &67.68 &22.68 &90.34 &67.57 &2.55\$ \\
Openai-ada-002 &65.34 &81.83 &71.76 &39.62 &76.79 &67.07 &1.66\$ \\
\toprule
\end{tabular}}
\caption{Domain-specific {\benchmark} results.}\label{tab: air_bench}
\end{table}

\begin{table}[!htp]\centering
\scriptsize
\resizebox{\columnwidth}{!}{%
\begin{tabular}{lcccr}\toprule
\textbf{Model} &\textbf{MSA-Cult} &\textbf{Egy-DIA} &\textbf{Mor-DIA} &\textbf{Avg.} \\\midrule
\textbf{\largem} &\textbf{82.19} &\textbf{83.55} &\textbf{65.35} &\textbf{77.03} \\
Cohere-v3.0 &\underline{81.86} &\underline{82.90} &\underline{65.23} &\underline{76.66} \\
OpenAI-3-large &81.49 &78.45 &64.90 &74.95 \\
Cohere-light-v3.0 &80.75 &64.82 &56.84 &67.47 \\
Me5-large &78.65 &61.34 &60.66 &66.88 \\
OpenAI-3-Small &74.55 &65.89 &54.13 &64.86 \\\midrule
\textbf{\smallm} &75.56 &60.35 &53.56 &63.16 \\
Me5-base &74.56 &56.34 &53.91 &61.60 \\
Me5-small &73.81 &53.56 &45.45 &57.61 \\
ATM-V2 &63.78 &23.45 &21.45 &36.23 \\
ARBERTv2 &9.34 &8.55 &4.67 &7.52 \\
MARBERT &2.73 &0.44 &0.19 &1.12 \\
\bottomrule
\end{tabular}}
\caption{Cultural {\benchmark} results.}\label{tab: ArabicCult}
\end{table}
\begin{table*}[!htp]\centering
\scriptsize
\resizebox{\textwidth}{!}{%
\begin{tabular}{lccccccccccc}\toprule
\textbf{Model} &\textbf{ArRTR} &\textbf{DOM-RTR} &\textbf{DIA-RTR} &\textbf{STS} &\textbf{PairCLF} &\textbf{CLF} &\textbf{RRK} &\textbf{CLK} &\textbf{BTM} &\textbf{Avg.} \\\midrule
\textbf{\ourmodel-Small} &15.12 &8.46 &7.52 &37.88 &62.87 &\underline{56.85} &62.21 &39.25 &1.99 &32.46 \\
\hspace{1em}+ Arabic &28.39 &39.34 &15.23 &41.49 &70.25 &51.89 &\underline{68.57} &39.12 &\underline{18.74} &41.45 \\
\hspace{1em}+ Synthetic-MSA &31.07 &\underline{40.45} &\underline{53.45} &\textbf{55.78} &\underline{74.23} &54.27 &\textbf{68.88} &\underline{39.43} &18.19 &\underline{48.42} \\
\hspace{1em}+ Synthetic-DOM &\textbf{32.01} &\textbf{49.02} &49.34 &\underline{52.90} &\textbf{75.45} &54.43 &67.45 &\textbf{40.56} &17.35 &\textbf{48.72} \\
\hspace{1em}+ Synthetic-DIA &\underline{31.20} &38.66 &\textbf{59.43} &51.23 &72.86 &\textbf{57.56} &66.67 &37.34 &\textbf{19.90} &48.32 \\\midrule
\textbf{\ourmodel-Large} &44.46 &64.52 &66.23 &48.63 &72.34 &50.43 &69.39 &38.28 &44.20 &55.39 \\
\hspace{1em}+ Arabic &54.53 &66.43 &70.34 &52.93 &75.24 &\underline{52.54} &\underline{70.49} &\underline{40.21} &48.35 &59.01 \\
\hspace{1em}+ Synthetic-MSA &56.34 &\underline{67.90} &\underline{72.89} &\textbf{57.89} &\textbf{76.90} &50.21 &\textbf{70.92} &\textbf{41.76} &\underline{62.34} &\underline{61.91} \\
\hspace{1em}+ Synthetic-DOM &\textbf{58.42} &\textbf{76.54} &71.65 &55.92 &75.19 &50.19 &70.21 &39.33 &51.23 &60.96 \\
\hspace{1em}+ Synthetic-DIA &\underline{57.09} &65.06 &\textbf{77.03} &\underline{56.90} &\underline{76.42} &\textbf{54.89} &69.32 &39.41 &\textbf{65.56} &\textbf{62.41} \\
\bottomrule
\end{tabular}}
\caption{The impact of Synthetic Data on {\ourmodel} performance. \textbf{ArRTR:} Arabic retrieval, \textbf{DOM-RTR:} Domain-specific retrieval, and \textbf{DIA-RTR}: Dialectal Retrieval. }\label{tab: synth_data}
\end{table*}

We evaluate the performance of our models, \smallm~and \largem, across the multiple proposed \benchmark~benchmarks and compare them with existing SoTA models, including MARBERT~\cite{abdul-mageed2020arbert}, ARBERTv2~\cite{elmadany2022orca}, CamelBERT~\cite{CAMeLBERT2021}, multilingual E5 models~\cite{wang2024multilingual}, and Arabic-triplet-Matryoshka-V2 (ATM-V2)~\cite{nacar2024enhancing}. Our evaluation results encompasses overall \benchmark~ (Table \ref{tab: results}), dialectal \benchmark~(Table \ref{tab: ArabicDialect}), domain-specific \benchmark~tasks (Table \ref{tab: air_bench}), and cultural \benchmark~(Table \ref{tab: ArabicCult}). In these tables, the tasks will be referred to as \textbf{RTR}: Retrieval, \textbf{STS}: Semantic Textual Similarity, \textbf{PairCLF}: Pair Classification, \textbf{CLF}: Classification, \textbf{CLR}: Clustering, \textbf{RRK}: Reranking, and \textbf{BTM}: BiText Mining. 

\noindent\textbf{\benchmark~Results.} Table \ref{tab: results} presents the overall results of our models on the \benchmark~benchmark. Our models demonstrate top-tier performance across a variety of NLP tasks. \smallm~achieves an average score of 57.33, surpassing its main competitors, Me5-base (55.29) and Me5-small (55.06), by a significant margin. This model performs exceptionally well in retrieval (58.42), classification (57.34), and pair classification (74.93), outperforming ATM-V2, which only scores 45.24 on average. Similarly, \largem~sets a new state-of-the-art performance with an average score of 62.45, beating Me5-large (61.65) and even the massive e5-mistral-7b model (59.00). The model excels particularly in retrieval (65.63), classification (54.89), and bitext mining (71.24), indicating its robustness across both cross-lingual and Arabic-centric tasks. These results validate our training strategy of using diverse training data covering multiple languages, where \ourmodel-Large outperforms its counterparts by more than five points in cross-lingual tasks such as bitext mining.


\noindent\textbf{Dialectal \benchmark~Results.} 
Table \ref{tab: ArabicDialect} shows the dialectal \benchmark results. \ourmodel-Small scores an average of 63.41, considerably higher than Me5-small (45.27) and AlcLaM (30.44), showing strong performance across retrieval (63.16) and classification (54.52). \ourmodel-Large achieves an impressive average score of 70.45, leading all tasks and outperforming the e5-mistral-7b model, which scores 60.81. The standout result is in bitext mining, which achieves 72.10, showcasing a substantial 14-point improvement over AlcLaM (59.38). Our models' significant advantage in dialectal retrieval and bitext mining is their unique training with a combination of synthetic and human-generated dialectal datasets, which is absent in many competitive models.

\noindent\textbf{Domain-Specific \benchmark~Results.}
As seen from Table~\ref{tab: air_bench}, \smallm~performs exceptionally well, with an average score of 70.86, surpassing OpenAI’s text-embedding-3-small model (68.65) and Cohere-light-v3.0 (67.57). Its best performance is in the legal domain, where it scores 78.86. \largem~sets a new standard in domain-specific tasks, scoring 82.49 on average, surpassing OpenAI’s text-embedding-3-large (82.20) and Cohere’s multilingual model (73.76). The model excels particularly in the news domain (90.42), medical (81.64), and Wikipedia (93.10), indicating its superior generalization across varied Arabic domains. Moreover, the cost-effectiveness of our models is evident: using \largem~costs only $0.75$ for 10k documents compared to $9.88$ for OpenAI’s model, making it a more efficient solution for large-scale deployments.

\noindent\textbf{Cultural ArabicMTEB Results.}
Cultural \benchmark~is designed to capture culturally sensitive aspects of the Arabic language, such as regional dialects, local idiomatic expressions, and culturally specific knowledge. We generated queries from country-specific Wikipedia articles, including questions about local cuisine, traditional practices, and historical events, which challenge the models to capture more than just linguistic information. For example, \largem~achieved the highest performance on tasks related to Egyptian cultural queries, outperforming other models on retrieval tasks by 1.5\%. However, we observed slightly lower performance on Moroccan dialect queries, where cultural nuances (such as regional vocabulary) presented a more significant challenge. 



\noindent\textbf{Synthetic Data Analysis.} \label{section: Synthetic}
We systematically analyze the impact of synthetic data on the performance of \smallm~and \largem~using different combinations of training datasets. Table \ref{tab: synth_data} presents the results for the base models, models trained with additional human-generated Arabic data, and models enhanced using synthetic subsets such as MSA, domain-specific, and dialectal data.
When comparing the initial \smallm~(average score of 32.46) to its version trained with synthetic MSA data, we observe a significant increase in average performance to \textbf{48.42}, representing an improvement of more than \textbf{16 points}. Similarly, \largem~benefits from a \textbf{6.52-point} boost in average performance (from 55.39 to 61.91) with the inclusion of synthetic MSA data.

\section{Conclusion} \label{section: conclusion}

In this paper, we introduced \smallm~and \largem, along with the comprehensive \benchmark~ benchmark for evaluating Arabic text embeddings. Our models demonstrate outstanding performance, benefiting from the strategic use of hard negatives and synthetic data in training. The evaluation across multiple benchmarks demonstrates that both \smallm~and \largem~set new standards in Arabic-centric NLP tasks. They outperform existing SoTA models in both cross-lingual and Arabic-specific tasks while being cost-effective and capable of understanding cultural context—making them ideal for real-world applications in diverse Arabic language settings. 

\section{Limitations}
While the development of the \ourmodel~models and the introduction of \benchmark~mark significant advancements in Arabic text embeddings, there are a number of limitations to consider. For example, although synthetic data significantly enhances model performance, it can introduce biases due to the reliance on specific patterns in the generated content. We ensured our synthetic data generation diversity by varying the data sources and generating dialectal data for multiple regions, including Egypt, Morocco, and the Gulf states, to mitigate this. We also analyzed our models by examining whether MSA data received higher accuracies in retrieval tasks in \autoref{tab: synth_data}. Further, our synthetic data generation pipeline was subjected to human verification for correctness and balance across cultural contexts.

\section{Ethical Statement}
The societal implications of deploying dialect-aware models, such as \ourmodel, require careful consideration. While these models can bridge gaps in NLP for Arabic-speaking regions, there is a risk of inadvertently reinforcing biases or language hierarchies, particularly in areas where particular dialects are stigmatized or underrepresented. For instance, users in communities with dialects associated with lower socioeconomic status may feel marginalized if their dialect is not adequately supported. To mitigate these concerns, we have prioritized the inclusion of low-resource dialects and ensured that our synthetic data generation pipeline accounts for dialectal diversity. Additionally, future versions of models should include further dialectal balancing, specifically focusing on underrepresented communities.

Importantly, all research and development activities for the \texttt{Swan} models and \texttt{ArabicMTEB} benchmark were conducted with a commitment to ethical standards. Data collection and usage adhered to privacy and confidentiality norms, ensuring no sensitive information was utilized without proper anonymization and consent. 
\section*{Acknowledgments}\label{sec:acknow}
We acknowledge support from Canada Research Chairs (CRC), the Natural Sciences and Engineering Research Council of Canada (NSERC; RGPIN-2018-04267), the Social Sciences and Humanities Research Council of Canada (SSHRC; 895-2020-1004; 895-2021-1008), Canadian Foundation for Innovation (CFI; 37771), Digital Research Alliance of Canada,\footnote{\href{https://alliancecan.ca}{https://alliancecan.ca}} and UBC Advanced Research Computing-Sockeye.\footnote{\href{https://arc.ubc.ca/ubc-arc-sockeye}{https://arc.ubc.ca/ubc-arc-sockeye}}

\bibliography{custom}

\appendix

\section{ArMistral Training}\label{Appendix: training_mistral}

ArMistral, is an autoregressive pretrained language model based on Mistral-7B.

\noindent\textbf{Pretraining data}
We further pretrain it on a large and diverse Arabic dataset, including all categories of Arabic, namely Classical Arabic (CA), Dialectal Arabic (DA), and MSA. This data is aggregated from various sources:  
AraNews\textsubscript{v2}~\cite{nagoudi2020machine}, El-Khair~\cite{elkhair-2016}, Gigaword,\footnote{\href{https://catalog.ldc.upenn.edu/LDC2009T30}{LDC Catalog Link}} OSCAR~\cite{suarez2019asynchronous}, OSIAN~\cite{zeroual2019osian}, 101 Billion arabic words \cite{aloui2024101}, Wikipedia Arabic, and Hindawi Books.\footnote{\href{https://www.hindawi.org/books/}{OpenITI corpus (v1.6).}} We also derived ArabicWeb22 (A) and (B) from the open source Arabic text~2022.\footnote{\href{https://data.baai.ac.cn/details/ArabicText-2022}{ArabicText-2022 data}} 
This pretraining dataset was cleaned, filtered and deduplicated using \citet{Gagan_PolyDeDupe_2023}. We have also ensured that the model is pretrained in multiple domains, enhancing its results as seen in Table \ref{tab: ArMistral}. 

\noindent\textbf{Instruction Finetuning.} To enhance the capabilities of our ArMistral, we instruct-tuning it on three datasets: Alpaca-GPT4, Evol-instruct, and ShareGPT extracted from MultilingualSIFT datasets~\cite{Chen_MultilingualSIFT_Multilingual_Supervised_2023}.

\noindent\textbf{Alignment Dataset.} We collected an alignment dataset from Quora and Mawdoo websites and then we took the gold answers as the choosen and we generated the rejected using AceGPT-7B \cite{huang-etal-2024-acegpt}.

\noindent\textbf{Results.} As seen from \autoref{tab: ArMistral}, Our ArMistral-Chat model outperforms all existing Arabic LLMs. 

\begin{table*}[h]\centering
\scriptsize
\resizebox{\textwidth}{!}{%
\begin{tabular}{lcccccccc}\toprule
\textbf{Model} &\textbf{ARC} &\textbf{Hellaswag} &\textbf{Exams} &\textbf{MMLU} &\textbf{Truthfulqa} &\textbf{ACVA} &\textbf{AlGhafa} &\textbf{Average} \\\midrule
\textbf{ArMistral-7B-Chat} &\underline{43.20} &\underline{55.53} &\underline{45.54} &\textbf{43.50} &\textbf{52.44} &\underline{77.06} &\textbf{35.57} &\textbf{50.41} \\
Jais-13b-chat &41.10 &\textbf{57.70} &\textbf{46.74} &\underline{42.80} &47.48 &72.56 &\underline{34.42} &\underline{48.97} \\
AceGPT-13B-chat &\textbf{43.80} &52.70 &42.09 &41.10 &\underline{49.96} &\textbf{78.42} &31.95 &48.57 \\
AceGPT-13B-base &39.90 &51.30 &39.48 &40.50 &46.73 &75.29 &30.37 &46.22 \\
AraLLama-7B-Chat &39.45 &50.23 &38.24 &41.03 &50.44 &70.45 &32.54 &46.05 \\
\textbf{ArMistral-7B-Base}&41.50 &52.50 &38.92 &37.50 &51.27 &69.64 &30.24 &45.94 \\
Jais-13b-base &39.60 &50.30 &39.29 &36.90 &50.59 &68.09 &30.07 &44.98 \\
AceGPT-7B-chat &38.50 &49.80 &37.62 &34.30 &49.85 &71.81 &31.83 &44.81 \\
AraLLama-7B-Base &38.40 &50.12 &38.43 &40.23 &45.32 &69.42 &31.52 &44.78 \\
AceGPT-7B-base &37.50 &48.90 &35.75 &29.70 &43.04 &68.96 &33.11 &42.42 \\
\bottomrule
\end{tabular}}
\caption{Comparison of ArMistral with other Arabic LLMs.}\label{tab: ArMistral}
\end{table*}

\section{Datasets overview}
\label{sec:appendix_datasets_benchmark}
The Table \ref{tab:datasets_overview} provides a comprehensive summary of the various datasets utilized in the study. It categorizes datasets based on their type, such as Reranking, Bitext Mining, Retrieval, Crosslingual Retrieval, STS, Pair Classification, Clustering, and Classification. Each entry specifies the dataset name, language, citation, and category, reflecting the diversity and scope of data sources for evaluating the model’s performance across different tasks and linguistic contexts.

\section{Polydedupe: versatile cleaning Pipeline} \label{sec:pipeline}
PolyDeDupe is a Python package designed for efficient and effective data deduplication across over 100 languages. It supports syntactic and semantic deduplication, making it a versatile tool for high-quality data preprocessing in NLP tasks. Key features include customizable Jaccard similarity thresholds, a performance speed twice that of other tools like SlimPajama, and support for deduplicating instruction tuning data. It can be easily installed via pip to deduplicate datasets, display original and filtered dataset sizes, and identify duplicate clusters. Supported languages span Western, Central, and Eastern European languages, Slavic languages using Cyrillic script, Greek, various Arabic and Devanagari script languages, and more.


\begin{table*}[!htp]\centering
\scriptsize
\resizebox{0.7\textwidth}{!}{%
\begin{tabular}{lrrrrrr}\toprule
\textbf{Task} &\textbf{Dataset} &\textbf{Type} &\textbf{Language} &\textbf{Citation} &\textbf{Size} \\\midrule
BitextMining &Darija &S2S &Moroccan Arabic Dialect to English &\cite{nagoudi2023dolphin} &2000 \\
BitextMining &Narabizi &S2S &Arabizi to French &\cite{nagoudi2023dolphin} &144 \\
BitextMining &Mt\_en2ar &S2S &English to MSA &\cite{nagoudi2023dolphin} &4000 \\
BitextMining &Mt\_fr2ar &S2S &French to MSA &\cite{nagoudi2023dolphin} &4000 \\
BitextMining &Mt\_es2ar &S2S &Spanish to MSA &\cite{nagoudi2023dolphin} &4000 \\
BitextMining &Mt\_ru2ar &S2S &Russian to MSA &\cite{nagoudi2023dolphin} &4000 \\
BitextMining &Cs\_dz\_fr &S2S &Algerian Arabic Dialect to French &\cite{nagoudi2023dolphin} &200 \\
BitextMining &Cs\_eg\_en &S2S &Egyptian Arabic Dialect to English &\cite{nagoudi2023dolphin} &200 \\
BitextMining &Cs\_jo\_en &S2S &Jordanian Arabic to English &\cite{nagoudi2023dolphin} &200 \\
BitextMining &Cs\_ma\_fr &S2S &Moroccan Arabic to French &\cite{nagoudi2023dolphin} &200 \\
BitextMining &Cs\_ps\_en &S2S &Palestinian Arabic to English &\cite{nagoudi2023dolphin} &200 \\
BitextMining &Cs\_ye\_en &S2S &Yemeni Arabic to English &\cite{nagoudi2023dolphin} &200 \\\midrule
Classification &MassiveIntent &S2S &Multilingual (Arabic subset) &\cite{fitzgerald2022massive} &100 \\
Classification &MassiveScenario &S2S &Multilingual (Arabic subset) &\cite{fitzgerald2022massive} &100 \\
Classification &OrcaSentiment &S2S &Arabic &\cite{elmadany2022orca} &5000 \\
Classification &OrcaDialect\_region &S2S &Arabic &\cite{elmadany2022orca} &5000 \\
Classification &OrcaDialect\_binary &S2S &Arabic &\cite{elmadany2022orca} &5000 \\
Classification &OrcaDialect\_country &S2S &Arabic &\cite{elmadany2022orca} &5000 \\
Classification &OrcaAns\_claim &S2S &Arabic &\cite{elmadany2022orca} &5000 \\
Classification &OrcaMachine\_generation &S2S &Arabic &\cite{elmadany2022orca} &5000 \\
Classification &OrcaAge &S2S &Arabic &\cite{elmadany2022orca} &5000 \\
Classification &OrcaGender &S2S &Arabic &\cite{elmadany2022orca} &5000 \\
Classification &OrcaAdult &S2S &Arabic &\cite{elmadany2022orca} &5000 \\
Classification &OrcaDangerous &S2S &Arabic &\cite{elmadany2022orca} &5000 \\
Classification &OrcaEmotion &S2S &Arabic &\cite{elmadany2022orca} &5000 \\
Classification &OrcaHate\_speech &S2S &Arabic &\cite{elmadany2022orca} &5000 \\
Classification &OrcaOffensive &S2S &Arabic &\cite{elmadany2022orca} &5000 \\
Classification &OrcaIrony &S2S &Arabic &\cite{elmadany2022orca} &5000 \\
Classification &OrcaSarcasm &S2S &Arabic &\cite{elmadany2022orca} &5000 \\
Classification &OrcaAbusive &S2S &Arabic &\cite{elmadany2022orca} &5000 \\\midrule
Clustering &Arabic\_news &P2P &Arabic &Our Paper &2500 \\
Clustering &Arabic\_topic &S2S &Arabic &Our Paper &30 \\
Clustering &Arabic\_baly\_stance &P2P &Arabic &\cite{elmadany2022orca} &1000 \\
Clustering &Arabic\_baly\_stance &S2S &Arabic &\cite{elmadany2022orca} &100 \\
PairClassification &Arabic\_xnli &S2S &Arabic &Our Paper &538 \\
PairClassification &Arabic\_sts &S2S &Arabic &Our Paper &1256 \\
PairClassification &Arabic\_mq2q &S2S &Arabic &Our Paper &244 \\\midrule
Reranking &Miracl\_ar &S2P &Multilingual (Arabic subset) &\cite{zhang-etal-2023-miracl} &750 \\
Reranking &Mmarco\_arabic &S2P &Arabic &Our Paper &3000 \\
Reranking &MedicalQA\_arabic &S2P &Arabic &Our Paper &4350 \\
Reranking &Mmarco\_en2ar &S2P &English to MSA &Our Paper &500 \\
Reranking &Mmarco\_ar2en &S2P &MSA to English &Our Paper &500 \\\midrule
Retrieval &MultiLongDoc &S2P &Multilingual (Arabic subset) &MDQA & \\
Retrieval &XPQA &S2S &Multilingual (Arabic subset) &XPQA & \\
Retrieval &Mintaka &S2S &Multilingual (Arabic subset) &Mintaka & \\
Retrieval &Lareqa &S2P &Arabic &\cite{nagoudi2023dolphin} &220 \\
Retrieval &Dawqs &S2S &Arabic &\cite{nagoudi2023dolphin} &318 \\
Retrieval &Exams &S2S &Arabic &\cite{nagoudi2023dolphin} &2600 \\
Retrieval &Mkqa &S2S &Arabic &\cite{nagoudi2023dolphin} &340 \\
Retrieval &Mlqa &S2S &Arabic &\cite{nagoudi2023dolphin} &517 \\
Retrieval &Arcd &S2S &Arabic &\cite{nagoudi2023dolphin} &693 \\
Retrieval &Tydiqa &S2S &Arabic &\cite{nagoudi2023dolphin} &5700 \\
Retrieval &Xsquad &S2S &Arabic &\cite{nagoudi2023dolphin} &5700 \\
Retrieval &Crosslingual\_ar2de &S2P &MSA to German &Our Paper &1831 \\
Retrieval &Crosslingual\_ar2en &S2P &MSA to English &Our Paper &1831 \\
Retrieval &Crosslingual\_ar2es &S2P &MSA to Spanish &Our Paper &1831 \\
Retrieval &Crosslingual\_ar2hi &S2P &MSA to Hindi &Our Paper &1831 \\
Retrieval &Crosslingual\_ar2vi &S2P &MSA to Vietnamese &Our Paper &1831 \\
Retrieval &Crosslingual\_ar2zh &S2P &MSA to Chinese &Our Paper &1831 \\
Retrieval &Crosslingual\_de2ar &S2P &German to MSA &Our Paper &1831 \\
Retrieval &Crosslingual\_en2ar &S2P &English to MSA &Our Paper &1831 \\
Retrieval &Crosslingual\_es2ar &S2P &Spanish to MSA &Our Paper &1831 \\
Retrieval &Crosslingual\_hi2ar &S2P &Hindi to MSA &Our Paper &1831 \\
Retrieval &Crosslingual\_vi2ar &S2P &Vietnamese to MSA &Our Paper &1831 \\
Retrieval &Crosslingual\_zh2ar &S2P &Chinese to MSA &Our Paper &1912 \\
Retrieval &MoroccoCultural &S2P &Arabic &Our Paper &100 \\
Retrieval &SyriaCultural &S2P &Arabic &Our Paper &100 \\
Retrieval &LibyaCultural &S2P &Arabic &Our Paper &100 \\
Retrieval &LebanonCultural &S2P &Arabic &Our Paper &100 \\
Retrieval &QatarCultural &S2P &Arabic &Our Paper &100 \\
Retrieval &SudanCultural &S2P &Arabic &Our Paper &100 \\
Retrieval &AlgeriaCultural &S2P &Arabic &Our Paper &100 \\
Retrieval &MauritaniaCultural &S2P &Arabic &Our Paper &100 \\
Retrieval &TunisiaCultural &S2P &Arabic &Our Paper &100 \\
Retrieval &IraqCultural &S2P &Arabic &Our Paper &100 \\
Retrieval &EgyptCultural &S2P &Arabic &Our Paper &100 \\
Retrieval &SomaliaCultural &S2P &Arabic &Our Paper &100 \\
Retrieval &UAE\_Cultural &S2P &Arabic &Our Paper &100 \\
Retrieval &OmanCultural &S2P &Arabic &Our Paper &100 \\
Retrieval &KuwaitCultural &S2P &Arabic &Our Paper &100 \\
Retrieval &BahrainCultural &S2P &Arabic &Our Paper &100 \\
Retrieval &Saudi\_ArabiaCultural &S2P &Arabic &Our Paper &100 \\
Retrieval &JordanCultural &S2P &Arabic &Our Paper &100 \\
Retrieval &PalestineCultural &S2P &Arabic &Our Paper &100 \\
Retrieval &YemenCultural &S2P &Arabic &Our Paper &100 \\
Retrieval &MoroccoDIA &S2P &Moroccan Arabic Dialect &Our Paper &100 \\
Retrieval &EgyptDIA &S2P &Egyptian Arabic Dialect &Our Paper &100 \\
Retrieval &NewsDomainSpecific &S2P &Arabic &Our Paper &1000 \\
Retrieval &LegalDomainSpecific &S2P &Arabic &Our Paper &1000 \\
Retrieval &MedicalDomainSpecific &S2P &Arabic &Our Paper &1000 \\
Retrieval &FinanceDomainSpecific &S2P &Arabic &Our Paper &1000 \\
Retrieval &WikipediaDomainSpecific &S2P &Arabic &Our Paper &1000 \\\midrule
STS &STS17 &S2S &Arabic &\cite{cer-etal-2017-semeval} &8060 \\
STS &STS22 &P2P &Arabic &\cite{semenov-etal-2023-findings} &500 \\
STS &Arabic\_sts &S2S &Arabic &Our Paper &750 \\
STS &Arabic\_stsb\_multi\_dialect &S2S &Arabic Dialectal &Our Paper &1500 \\
STS &Arabic\_sts &P2P &Arabic &Our Paper &500 \\
\bottomrule
\end{tabular}}
\caption{Overview of \benchmark~datasets. \textbf{S2S:} Sentence to Sentence. \textbf{S2P:} Sentence to Paragraph. \textbf{P2P:} Paragraph to Paragraph.}
\label{tab:datasets_overview}
\end{table*}

\section{Prompts for evaluation}
Table~\ref{tab: defs} provides an overview of the prompts used for evaluating various tasks. It includes instructions for Reranking, Bitext Mining, Retrieval, Crosslingual Retrieval, Semantic Textual Similarity (STS), Pair Classification, Clustering, and Classification. Each entry outlines the specific task and the corresponding instruction used to guide the model’s evaluation process.
\begin{table*}[t]\centering
\scriptsize
\resizebox{\textwidth}{!}{%
\begin{tabular}{ll}\toprule
\textbf{Task} &\textbf{Instructions} \\\midrule
Reranking &Given an Arabic search query, retrieve web passages that answer the question in \{Lang\}. Query:\{query\}. \\
BitextMining &Retrieve parallel sentences in \{Lang\}. \\
Retrieval &Given an Arabic search query, retrieve web passages that answer the question. Query:\{query\}.  \\
Crosslingual Retrieval &Given an Arabic search query, retrieve web passages that answer the question in \{Lang\}. Query:\{query\}. \\ 
STS &Retrieve semantically similar text. Text: \{text\}.  \\
Pair Classification &Retrieve texts that are semantically similar to the given text. Text: \{text\}. \\
Clustering &Identify the topic or theme of the given news article. Article:\{article\}.  \\
Classification &Classify the text into the given categories \{options\}. \\
\bottomrule
\end{tabular}}
\caption{Prompts used for evaluation.}\label{tab: defs}
\end{table*}

\section{Full Leaderboard}
Table~\ref{tab: big results} presents the performance comparison of various models on different tasks within the \benchmark~benchmark. It includes metrics for Retrieval, Semantic Textual Similarity (STS), Pair Classification (PairCLF), Classification (CLF), Re-ranking, Clustering, and Bitext Mining (BTM). The table lists each model, its dimensionality, and the scores for each task, along with an overall average score. The results highlight the strengths and weaknesses of each model across a range of tasks, providing a comprehensive overview of their performance.
\begin{table*}[!htp]\centering
\scriptsize
\resizebox{\textwidth}{!}{%
\begin{tabular}{lrcccccccc}\toprule
\textbf{Model} &\textbf{Dim.} &\textbf{Retrieval} &\textbf{STS} &\textbf{PairCLF} &\textbf{CLF} &\textbf{Re-rank} &\textbf{Cluster} &\textbf{BTM} &\textbf{Avg} \\\cmidrule{1-10}
\multicolumn{2}{c}{\textbf{Number of datasets}} &\textbf{23} &\textbf{5} &\textbf{3} &\textbf{18} &\textbf{5} &\textbf{4} &\textbf{12} &\textbf{70} \\\midrule
\textbf{\largem} &4096 &\textbf{65.63} &\underline{59.10} &\textbf{75.62} &52.55 &69.42 &41.24 &\textbf{71.24} &\textbf{62.11} \\
multilingual-e5-lg &1024 &\underline{64.01} &\textbf{59.45} &\underline{75.06} &53.43 &\textbf{70.79} &\underline{42.49} &66.33 &\underline{61.65} \\
e5-mistral-7b-inst &4096 &56.34 &57.02 &70.24 &53.21 &66.24 &39.44 &\underline{70.50} &59.00 \\
\textbf{\smallm} &768 &58.42 &58.44 &74.93 &\textbf{57.34} &68.43 &40.43 &42.45 &57.21 \\
multiling-e5-b &768 &56.91 &57.99 &74.30 &52.30 &\underline{69.07} &\textbf{42.56} &33.90 &55.29 \\
multiling-e5-s &384 &55.14 &56.73 &73.97 &50.85 &67.92 &42.37 &38.47 &55.06 \\
LaBSE &768 &34.98 &54.15 &70.60 &49.57 &62.17 &41.42 &33.28 &49.45 \\
text2vec-base &384 &27.69 &59.37 &71.41 &47.94 &57.76 &37.26 &38.32 &48.54 \\
ARBERTv2 &768 &15.12 &37.88 &62.87 &\underline{56.85} &62.21 &39.25 &1.99 &39.45 \\
CamelBERT-msa &768 &9.21 &47.69 &67.43 &55.77 &60.20 &39.89 &1.85 &40.29 \\
arabertv02-large &1024 &7.34 &34.26 &63.63 &54.32 &56.71 &37.26 &10.97 &37.78 \\
arabertv02-base &768 &8.62 &39.77 &66.30 &55.77 &60.03 &41.74 &0.70 &38.99 \\
CamelBERT-mix &768 &7.19 &46.47 &67.23 &56.68 &57.50 &38.72 &0.41 &39.17 \\
MARBERTv2 &768 &5.88 &45.21 &70.89 &54.89 &58.64 &40.81 &0.45 &39.54 \\
ARBERT &768 &8.07 &29.89 &61.86 &56.92 &61.09 &37.10 &2.28 &36.74 \\
CamelBERT-da &768 &4.07 &41.05 &65.82 &53.75 &54.44 &37.63 &0.31 &36.72 \\
MARBERT &768 &2.22 &40.62 &66.46 &54.35 &53.09 &36.33 &0.40 &36.21 \\
CamelBERT-ca &768 &2.74 &36.49 &62.26 &46.26 &51.34 &35.77 &0.09 &33.56 \\
\bottomrule
\end{tabular}}
\caption{\benchmark~ Results.}\label{tab: big results}
\end{table*}

\section{Inference Latency.}\begin{figure}[t]
\includegraphics[width=.47\textwidth]{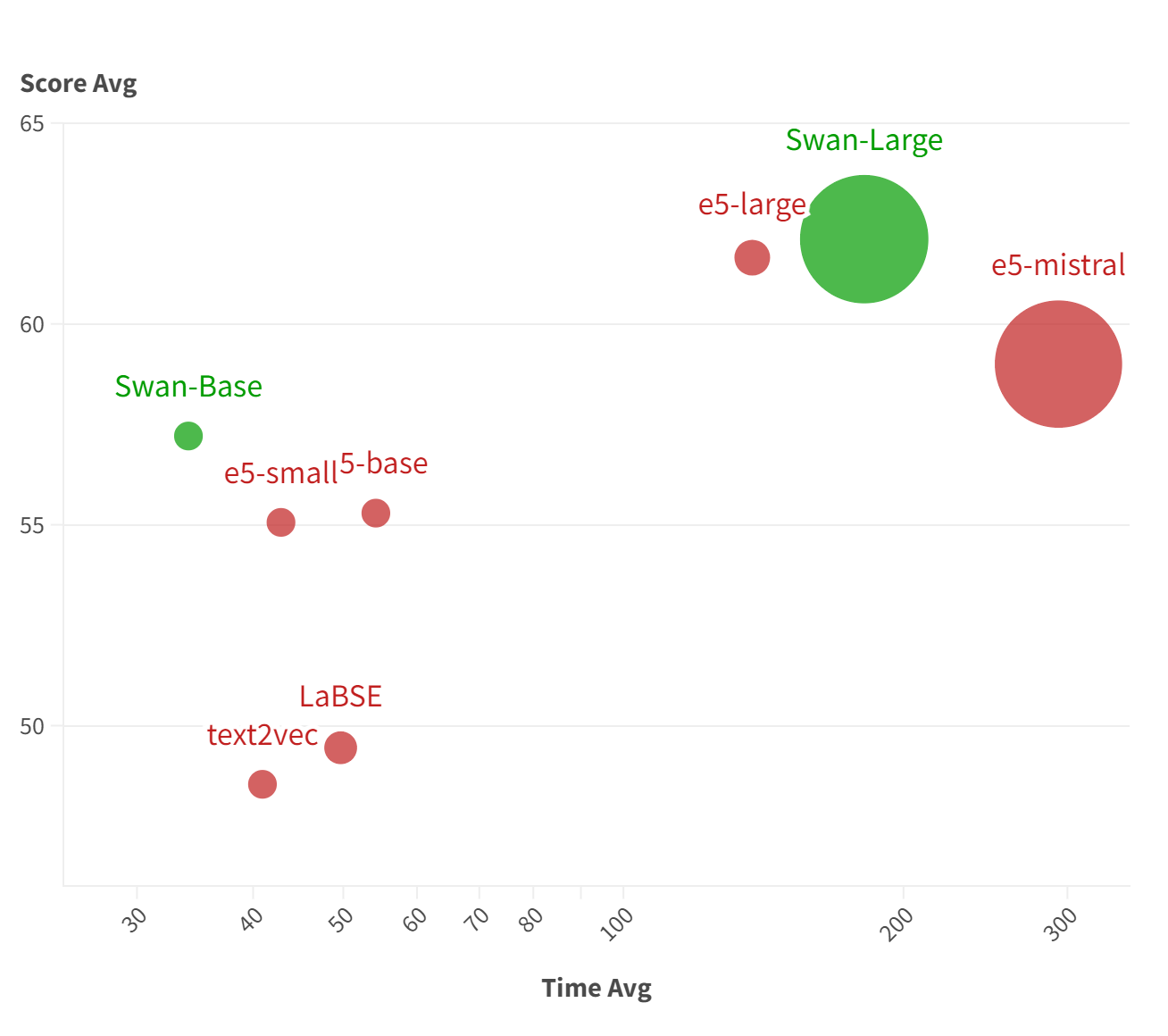}
\centering
\caption{Latency vs Performance.
}
\label{fig:Latency}
\end{figure} Inference latency is very critical in deploying machine learning models, especially in real-time applications with crucial response time. It refers to the time taken by a model to predict received input. In the context of text embedding models such as \smallm~and \largem, lower latency is particularly valuable for user-facing services that rely on fast processing of natural language input, such as chatbots and search engines.  From \autoref{fig:Latency}, we find that \largem, despite its larger size indicated by a larger bubble, has optimized inference times due to architectural efficiencies, and \smallm~strikes the perfect balance between size, performance, and latency. We compare the performance of the models from \autoref{tab: results}.

\section{STS Dataset Creation:} \label{sec:sts} The Arabic Semantic Textual Similarity (Arabic-STS) datasets was developed to facilitate research in semantic similarity for the Arabic language. The dataset is derived from the Arabic Billion Words \cite{aloui2024101billionarabicwords} corpus, which serves as a foundation for extracting a diverse collection of sentence pairs. Each pair is annotated with a similarity score that captures the degree of semantic equivalence between the sentences. The dataset generation process was guided by the capabilities of the GPT-4 model developed by OpenAI, ensuring that the resulting sentence pairs are of high quality and reflect nuanced linguistic characteristics. The creation involved several steps, including selecting representative sentences from the corpus, generating semantically varied sentence pairs, and annotating similarity scores using both automated methods and human reviewers to maintain consistency and reliability.

\section{Country level Cultural Evaluation}\label{sec: country_cul}

\begin{table*}[!htp]\centering
\scriptsize
\resizebox{\textwidth}{!}{%
\begin{tabular}{lrrrrrrrrrrrr}\toprule
\textbf{Model} &\textbf{\texttt{Swan-lg}} &\textbf{Me5-lg} &\textbf{Coh-lt-v3.0} &\textbf{\texttt{Swan-s}} &\textbf{OpenAI-3-lg} &\textbf{Coh-v3.0} &\textbf{Me5-s} &\textbf{Me5-b} &\textbf{ATM-V2} &\textbf{ARBERTv2} &\textbf{MARBERT} \\\midrule
Algeria &89.34 &93.34 &89.44 &90.45 &86.95 &88.99 &91.23 &90.66 &84.99 &18.27 &1.50 \\
Bahrain &93.71 &93.77 &93.52 &86.48 &91.98 &92.40 &93.08 &89.04 &90.49 &27.48 &5.74 \\
Egypt &98.34 &94.58 &91.37 &95.66 &91.45 &87.81 &93.02 &91.65 &88.45 &11.54 &1.63 \\
Iraq &92.45 &90.90 &86.98 &88.34 &92.43 &87.83 &89.02 &90.78 &81.22 &17.34 &1.92 \\
Jordan &92.34 &92.79 &90.07 &89.70 &94.56 &91.18 &93.67 &92.25 &87.95 &27.46 &4.50 \\
Kuwait &93.45 &96.34 &96.10 &90.44 &88.53 &92.51 &96.17 &94.94 &89.97 &36.67 &4.92 \\
Lebanon &95.66 &93.05 &92.38 &90.45 &90.23 &91.04 &91.92 &92.85 &87.14 &22.55 &1.82 \\
Libya &89.56 &88.43 &87.27 &85.45 &89.66 &85.75 &87.21 &85.32 &79.95 &28.88 &2.46 \\
Mauritania &92.44 &92.92 &92.61 &89.45 &90.31 &92.05 &20.99 &3.32 &0.63 &0.50 &0.00 \\
Morocco &90.34 &85.49 &83.19 &86.34 &83.56 &85.47 &81.73 &86.59 &4.75 &0.32 &0.00 \\
Oman &94.45 &94.26 &92.37 &91.98 &92.45 &92.61 &93.00 &93.04 &84.21 &11.24 &3.43 \\
Palestine &90.45 &90.67 &87.50 &91.18 &87.45 &83.33 &85.22 &86.49 &77.83 &27.25 &3.63 \\
Qatar &98.79 &93.44 &91.80 &92.35 &95.66 &89.98 &91.20 &90.49 &85.50 &29.15 &7.00 \\
Saudi\_Arabia &95.34 &93.49 &92.98 &91.47 &90.45 &92.12 &92.72 &91.47 &86.48 &25.06 &2.50 \\
Somalia &90.23 &94.78 &93.67 &88.34 &89.55 &92.30 &21.25 &2.50 &20.81 &2.62 &0.00 \\
Sudan &92.36 &91.99 &86.90 &90.89 &91.45 &90.72 &89.49 &87.60 &82.47 &24.51 &2.50 \\
Syria &91.46 &91.83 &90.56 &90.45 &90.56 &86.97 &88.69 &88.75 &87.45 &13.81 &3.63 \\
Tunisia &94.57 &94.64 &93.46 &95.54 &85.34 &90.92 &93.79 &92.04 &84.40 &25.04 &4.15 \\
UAE &96.09 &95.14 &93.41 &94.12 &97.66 &93.53 &94.45 &91.56 &91.79 &31.92 &2.00 \\
Yemen &92.34 &91.24 &89.40 &92.12 &89.54 &89.70 &88.25 &89.89 &83.08 &5.29 &1.29 \\\midrule
Avg. &93.19 &92.65 &90.75 &90.56 &90.49 &89.86 &83.81 &81.56 &73.98 &19.34 &2.73 \\
\bottomrule
\end{tabular}}
\caption{Country-level cultural evaluation.}\label{tab: countries}
\end{table*}

\end{document}